\documentclass[10pt,twocolumn,letterpaper]{article}

\usepackage[pagenumbers]{cvpr} 

\usepackage[dvipsnames]{xcolor}

\usepackage[utf8]{inputenc} 
\usepackage[T1]{fontenc}    
\usepackage{url}            
\usepackage{booktabs}       
\usepackage{amsfonts}       
\usepackage{nicefrac}       
\usepackage{microtype}      

\usepackage{graphicx}
\usepackage{amsmath}
\usepackage{amssymb}

\usepackage{diagbox}
\usepackage{multicol}
\usepackage{enumerate}
\usepackage{times}
\usepackage{epsfig}
\usepackage{threeparttable}
\usepackage{enumitem}
\usepackage{multirow}
\usepackage{color}
\usepackage{array}
\usepackage{setspace}
\usepackage{makecell}
 \usepackage{indentfirst}

\definecolor{cvprblue}{rgb}{0.21,0.49,0.74}
\usepackage[pagebackref,breaklinks,colorlinks,citecolor=cvprblue]{hyperref}

\def\paperID{2089} 
\def\confName{CVPR}
\def\confYear{2024}

\title{Bring Event into RGB and LiDAR: \\ Hierarchical Visual-Motion Fusion for Scene Flow}

\author{Hanyu Zhou\textsuperscript{\rm 1}, Yi Chang\textsuperscript{\rm 1}\thanks{Corresponding author.}, Zhiwei Shi\textsuperscript{\rm 1}, Luxin Yan\textsuperscript{\rm 1}\\
  \textsuperscript{\rm 1} National Key Lab of Multispectral Information Intelligent Processing Technology,\\School of Artificial Intelligence and Automation, Huazhong University of Science and Technology\\
  {\tt\small {\{hyzhou, yichang, shizhiwei, yanluxin\}}@hust.edu.cn}
 }

\usepackage[sort&compress]{natbib}
\begin{document}
\maketitle

\begin{abstract}

Single RGB or LiDAR is the mainstream sensor for the challenging scene flow, which relies heavily on visual features to match motion features. Compared with single modality, existing methods adopt a fusion strategy to directly fuse the cross-modal complementary knowledge in motion space. However, these direct fusion methods may suffer the modality gap due to the visual intrinsic \textbf{heterogeneous} nature between RGB and LiDAR, thus deteriorating motion features. We discover that event has the \textbf{homogeneous} nature with RGB and LiDAR in both visual and motion spaces. In this work, we bring the event as a bridge between RGB and LiDAR, and propose a novel hierarchical visual-motion fusion framework for scene flow, which explores a homogeneous space to fuse the cross-modal complementary knowledge for physical interpretation. In visual fusion, we discover that event has a complementarity (relative v.s. absolute) in luminance space with RGB for high dynamic imaging, and has a complementarity (local boundary v.s. global shape) in scene structure space with LiDAR for structure integrity. In motion fusion, we figure out that RGB, event and LiDAR are complementary (spatial-dense, temporal-dense v.s. spatiotemporal-sparse) to each other in correlation space, which motivates us to fuse their motion correlations for motion continuity. The proposed hierarchical fusion can explicitly fuse the multimodal knowledge to progressively improve scene flow from visual space to motion space. Extensive experiments have been performed to verify the superiority of the proposed method.

\end{abstract}

\begin{figure}
  \setlength{\abovecaptionskip}{5pt}
  \setlength{\belowcaptionskip}{-5pt}
  \centering
   \includegraphics[width=0.99\linewidth]{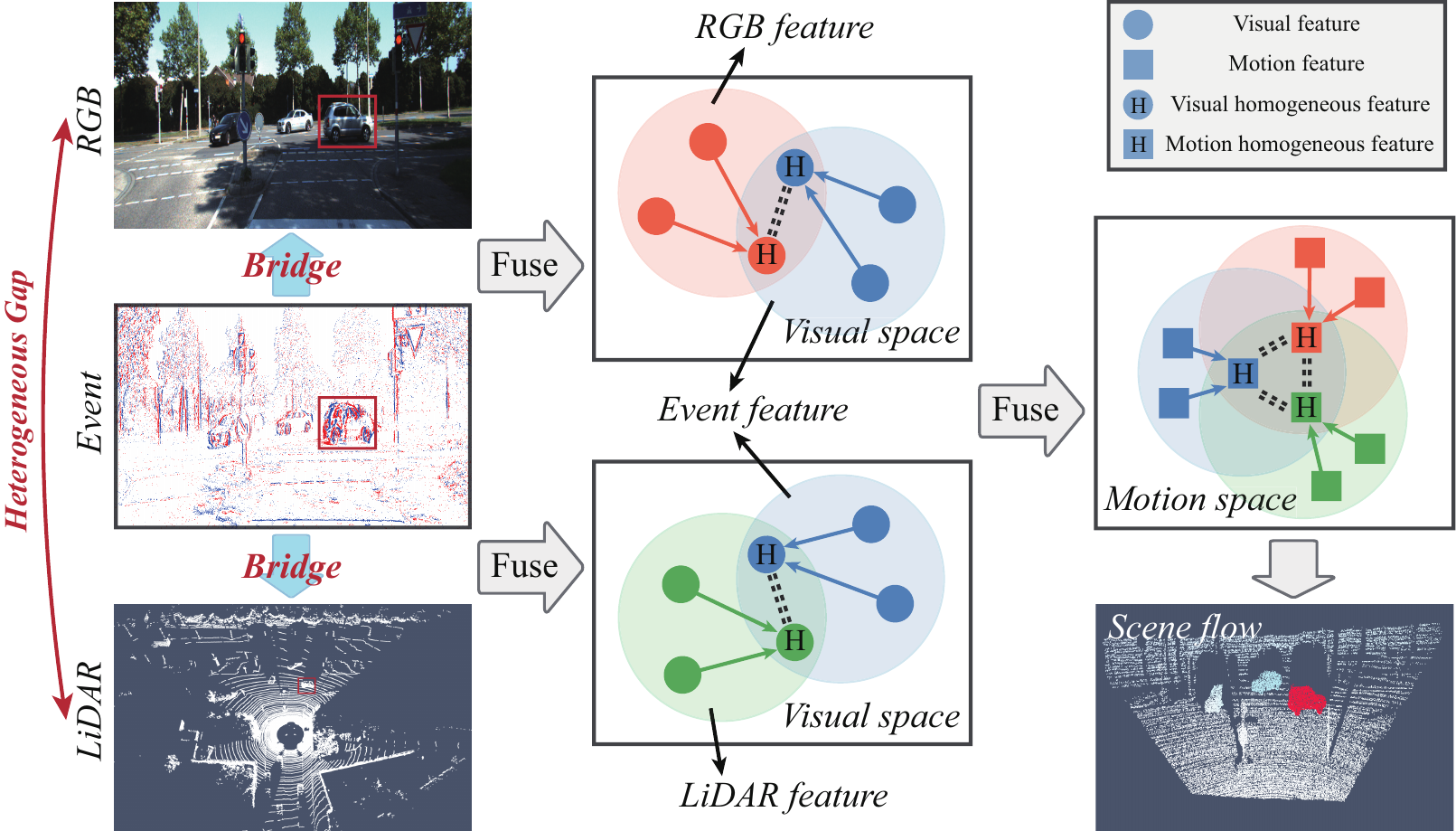}
   \caption{Illustration of the main idea. There exists a large modality gap due to the visual intrinsic heterogeneous nature between RGB and LiDAR, thus deteriorating the motion features. We discover that the event has the homogeneous nature with RGB and LiDAR in both visual and motion spaces. In this work, we bring the event as a bridge between RGB and LiDAR, and propose a novel hierarchical visual-motion fusion framework for scene flow, which explores a homogeneous feature space to explicitly fuse the cross-modal complementary knowledge for physical interpretation.
   }
   \label{Main_Idea}

\end{figure}

\section{Introduction}
\label{sec:intro}
Scene flow aims to model the correspondence between adjacent visual RGB or LiDAR features to estimate 3D motion features, which has been applied in various vision tasks, \emph{e.g.}, 3D object detection \cite{duffhauss2020pillarflownet} and motion segmentation \cite{baur2021slim}. The key to scene flow estimation is the highly discriminative visual feature for the valid motion feature matching.

Existing scene flow methods can be divided into two categories: unimodal direct learning methods \cite{behl2017bounding, ma2019deep, hur2021self, teed2021raft, wei2021pv} and multimodal motion fusion methods \cite{rishav2020deepLiDARflow, liu2022camliflow}. Unimodal direct learning methods mainly resort to the ideal visual feature to directly learn the motion feature from single RGB or LiDAR modality. Multimodal motion fusion methods further exploit the complementary knowledge between RGB and LiDAR to fuse their features in the motion space. For example, Liu \emph{et al.} \cite{liu2022camliflow} fused dense RGB motion features and sparse LiDAR motion features for scene flow.
However, these methods neglect the visual intrinsic heterogeneous nature, \emph{e.g.}, degraded visual RGB features with weakened texture under low light due to low dynamic range and limited visual LiDAR features with incomplete contour due to non-uniform laser beam. These can enlarge the modality gap and deteriorate the motion features. Therefore, \emph{we suggest that introducing an auxiliary modality as the bridge to assistantly enhance the RGB and LiDAR features in both visual and motion spaces.}

To our investigation, event camera \cite{Gallego2019EventBasedVA} is a neuromorphic sensor, which can sense the dynamic scene boundary structure with high dynamic range and high temporal resolution. In visual space, RGB records the absolute value of luminance while the event triggers the relative change of luminance, and LiDAR reflects the global shape of the scene while the event senses the local boundary. In motion space, we transform the visual features of these modalities into correlation space, and we figure out that RGB provides the spatial-dense 2D features, event provides the temporal-dense 2D features, and LiDAR provides the spatiotemporal-sparse 3D features. Therefore, the event has the complementarity with RGB and LiDAR in visual homogeneous (\emph{i.e.}, luminance and structure) and motion homogeneous (\emph{i.e.}, correlation) spaces.

In this work, we bring the auxiliary event as a bridge between RGB and LiDAR in Fig. \ref{Main_Idea}, and propose a novel hierarchical visual-motion fusion framework for scene flow (VisMoFlow), which \emph{focuses on fusing the cross-modal complementary knowledge in the homogeneous space.} In visual luminance fusion, we fuse the relative luminance of event and the absolute luminance of RGB for high dynamic imaging with the theoretically derived spatiotemporal gradient consistency. In visual structure fusion, we fuse the local boundary of event into the global shape of LiDAR for physical structure integrity using the self-similarity clustering strategy. In motion correlation fusion, we map event, RGB and LiDAR into the same motion manifold, and fuse the x, y-axis spatial-dense correlation features of RGB, the x, y-axis temporal-dense correlation features of event and the x, y, z-axis sparse correlation features of LiDAR for 3D motion spatiotemporal continuity by aligning the motion distributions between various modalities. The proposed hierarchical visual-motion fusion can explicitly learn the multimodal complementary knowledge from visual to motion space for scene flow. Overall, the main contributions are as follows:
\begin{itemize}[leftmargin=10pt]
\item We bring the auxiliary event as a bridge between RGB and LiDAR modalities, and propose a novel hierarchical visual-motion fusion framework for scene flow, which can explicitly fuse the multimodal knowledge to progressively improve scene flow from visual space to motion space.

\item We reveal that RGB, event and LiDAR modalities have the complementary knowledge in both visual homogeneous (\emph{i.e.}, luminance and structure) and motion homogeneous (\emph{i.e.}, correlation) spaces, which can make the entire multimodal fusion process more physically interpretable.

\item We conduct extensive experiments on daytime and nighttime scenes, which verify that the VisMoFlow achieves state-of-the-art performance for all-day scene flow.
\end{itemize}

\section{Related Work}
\label{sec:related_work}
\noindent
\textbf{Unimodal Scene Flow Estimation.}
Since scene flow is the 3D version of optical flow, the previous scene flow methods \cite{ilg2017flownet, behl2017bounding, ma2019deep, ranjan2019competitive, hur2021self, teed2021raft, liu2019flownet3d, wei2021pv} mainly directly follow the optical flow framework \cite{teed2020raft, jiang2021transformer}. They take monocular \cite{brickwedde2019mono, guizilini2022learning, hur2021self} or stereo sequences \cite{ranjan2019competitive, teed2021raft} as input, and calculate cost volume with warp operation through feature pyramid \cite{sun2018pwc} or recurrent network \cite{teed2020raft} for 3D motion.
However, since 2D RGB images lack the intrinsic 3D geometry property, LiDAR becomes a trend in scene flow estimation due to its 3D scene perception capability. Most LiDAR-based scene flow methods \cite{liu2019flownet3d, wei2021pv, shen2023self, lang2023scoop} also follow the architecture of the optical flow framework, while the difference is that they use MLP or voxel to represent the point cloud features. However, due to the potential incomplete physical contour of LiDAR, these single LiDAR-based scene flow methods are still limited. Instead, there is a strong complementarity between the two modalities, which inspires us to fully exploit the cross-modal complementary knowledge for scene flow.

\noindent
\textbf{Multimodal Scene Flow Estimation.}
As unimodal scene flow methods provide the partial information, multimodal fusion strategies \cite{ding2023hidden, rishav2020deepLiDARflow, liu2022camliflow, wan2023rpeflow} have gradually received attention.
Liu \emph{et al.} \cite{liu2022camliflow} proposed a bidirectional 2D-3D motion feature fusion framework, which can exploit the complementary features between RGB and LiDAR. Wan \emph{et al.} \cite{wan2023rpeflow} introduced the event camera to enhance the RGB and LiDAR features in motion space via a cross-attention mechanism.
However, these methods may suffer the potential degraded visual RGB features with weakened texture and limited visual LiDAR features with incomplete contour, thus leading to a large modality gap and deteriorating the valid motion features. In this work, we will introduce the auxiliary event as a bridge to hierarchically enhance the RGB and LiDAR features in both visual and motion spaces.

\noindent
\textbf{Multimodal Fusion.} Multimodal fusion focuses on exploiting the inter-modality complementarity for the target task.
To our knowledge, most existing multimodal fusion methods learn the mutual knowledge between different modalities via cross-attention mechanism \cite{hori2017attention, wei2020multi, sun2022event, cai2023objectfusion} or similarity representation learning \cite{bengio2013representation, hu2017learning, zhang2021rgb, lu2023see}.
For example, Cai \emph{et al.} \cite{cai2023objectfusion} proposed to utilize transformer \cite{vaswani2017attention} to directly fuse the proposal features of RGB images and LiDAR for 3D object detection. However, these methods neglect the modality gap due to the intrinsic heterogeneous nature of feature representation between various modalities, limiting the complementary knowledge fusion. To alleviate this issue, we introduce an auxiliary modality as the bridge to mitigate the modality gap, and explore the homogeneous space to explicitly fuse the cross-modal complementary knowledge for scene flow.

\begin{figure*}
  \setlength{\abovecaptionskip}{5pt}
  \setlength{\belowcaptionskip}{-5pt}
  \centering
   \includegraphics[width=1.0\linewidth]{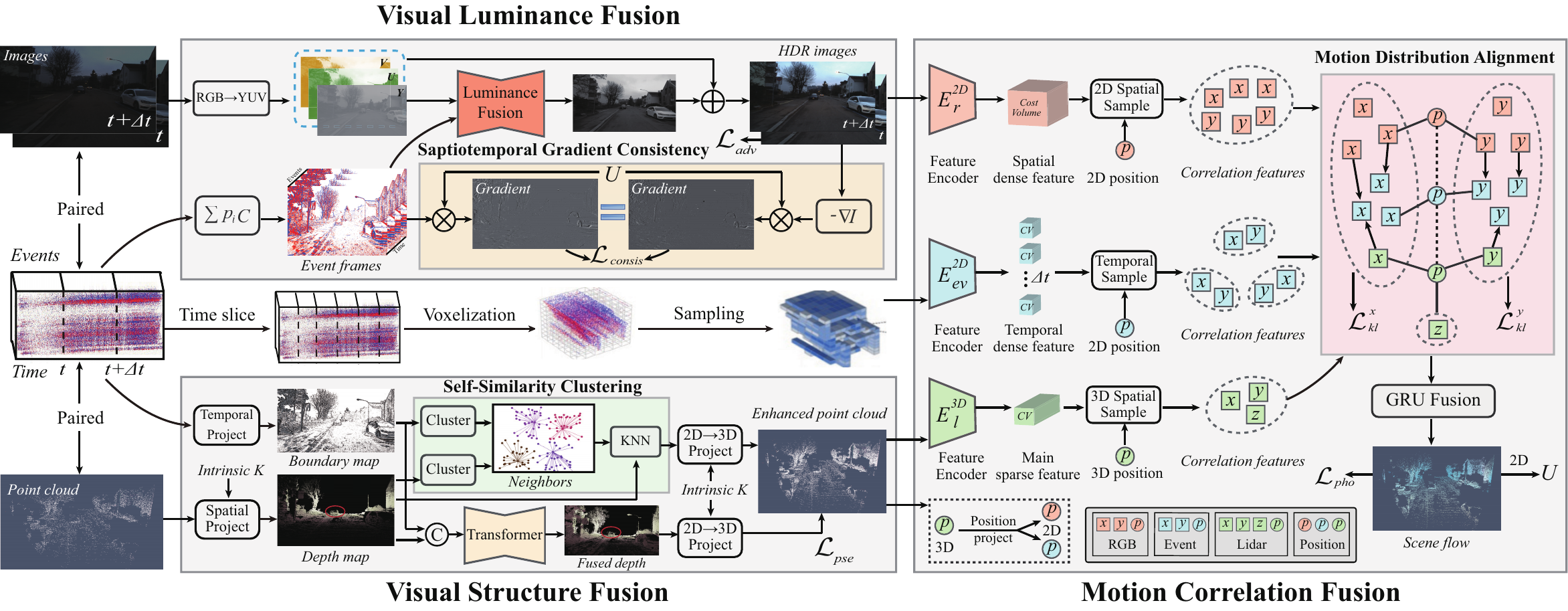}
   \caption{The architecture of the VisMoFlow mainly contains visual luminance fusion, visual structure fusion and motion correlation fusion. In visual luminance fusion, we fuse the relative luminance of event and the absolute luminance of RGB for high dynamic imaging. In visual structure fusion, we fuse the local boundary structure of event and the global shape structure of LiDAR for structure integrity. In motion correlation fusion, we fuse the spatiotemporal complementary correlation knowledge of the three modalities for 3D motion continuity.
   }
   \label{Framework}

\end{figure*}

\section{Hierarchical Visual-Motion Fusion}
\label{sec:method}
\subsection{Overall Framework}
Multimodal scene flow framework aims to fuse the cross-modal complementary knowledge for enhancing the discriminative visual and motion features that scene flow relies on. However, there exists a modality gap due to the visual intrinsic heterogeneous nature between RGB and LiDAR, deteriorating motion features. To this end, we introduce the auxiliary event as a bridge to mitigate the gap between RGB and LiDAR, and propose a novel multimodal hierarchical visual-motion fusion framework for scene flow in Fig. \ref{Framework}. The whole architecture looks complicated but is simply regarded as a task that builds a homogeneous space to fuse the cross-modal complementary knowledge for physical interpretation. In visual luminance fusion, we transform the event and RGB into the luminance space, and fuse the complementary (relative v.s. absolute) knowledge for high dynamic imaging under the constraint of the similar spatiotemporal gradient. In visual structure fusion, we represent the event and LiDAR into the spatial structure space, and fuse the complementary (local boundary v.s. global shape) knowledge for physical structure integrity using the self-similarity clustering strategy. In motion correlation fusion, we map the visual features of RGB, event and LiDAR to the same correlation space, and fuse the complementary (x, y-axis spatial-dense correlation, x, y-axis temporal-dense correlation and x, y, z-axis sparse correlation) knowledge for 3D motion spatiotemporal continuity via motion distribution alignment. Under the unified framework, the proposed method can explicitly fuse the cross-modal complementary knowledge to progressively improve scene flow from visual space to motion space. Next, we will describe the hierarchical visual-motion fusion framework in the form of ``what to fuse'' and ``how to fuse''.

\subsection{Visual Luminance Fusion}
Under oversaturated and low-light conditions, the RGB image may loss the texture, thus mismatching the motion features. The main reason is that RGB imaging mechanism limits the dynamic range. Therefore, we introduce event camera with the advantage of high dynamic range to assistantly enhance the visual RGB features.

\noindent
\textbf{Relative \emph{v.s.} Absolute Luminance.}
Event camera can asynchronously capture the events triggered by the brightness change, recording the relative luminance, while RGB camera records the absolute luminance via global scan. Motivated by this, we argue that event and RGB share the homogeneous luminance space, where the relative value of event and the absolute value of RGB are complementary to each other. Specifically, given the RGB image $I$ and the corresponding event stream, we transform the RGB image into YUV color space $[I^Y, I^U, I^V]$, where $I^Y$ denotes the luminance, $I^U$, $I^V$ is the chrominance information. We also accumulate the event stream to generate the intensity frame:
\begin{equation}
  \setlength\abovedisplayskip{1pt}
  \setlength\belowdisplayskip{1pt}
\begin{aligned}
I^{X} = \sum\nolimits {p_i}C,
  \label{eq:intensity_frame}
\end{aligned}
\end{equation}
where $p_i$ denotes the event polarity. $C$ is the event trigger threshold. Therefore, the absolute value $I^Y$ from the RGB image and the relative value $I^X$ from the event are complementary in the homogeneous luminance space. This makes us naturally transform RGB and event into a unified luminance space to achieve complementary knowledge fusion.

\noindent
\textbf{Spatiotemporal Gradient Consistency.}
We set various weights to fuse the absolute luminance of the RGB and the relative luminance of the event as follows:
\begin{equation}
  \setlength\abovedisplayskip{1pt}
  \setlength\belowdisplayskip{1pt}
\begin{aligned}
  H^Y = \mathcal{W}(I^Y, I^X) = (\omega_{y} I^{Y} + \omega_{x} I^{X})/(\omega_{y}+\omega_{x}),
  \label{eq:fusion_process}
\end{aligned}
\end{equation}
where $\omega_{x}$, $\omega_{y}$ denote the weights of event and RGB, respectively. The fused luminance image $H^Y$ has a high dynamic range, but without color information. The color can be compensated from U, V channels $I^U$, $I^V$ of the input RGB images, namely $H^{YUV} = \mathcal{C}(H^Y, I^U, I^V)$. We then merge this equal and Eq. \ref{eq:fusion_process} as follows:
\begin{equation}
  \setlength\abovedisplayskip{0pt}
  \setlength\belowdisplayskip{0pt}
\begin{aligned}
  H = \mathcal{C}(\mathcal{W}(I^Y, I^X), I^U, I^V).
  \label{eq:fusion_network}
\end{aligned}
\end{equation}

The fused YUV image $H$ should be further inversely transformed to the RGB color space. We follow \cite{10036136} to utilize the luminance fusion network to model the Eq. \ref{eq:fusion_network}. We take the adjacent RGB images [$I_{t}, I_{t+ \Delta t}$] at the timestamps $t$ and $t+\Delta t$, and the corresponding event stream as input, and output the fused RGB images [$\tilde{I}_{t}, \tilde{I}_{t+ \Delta t}$]. Note that, the fused images have two limitations: slightly unnatural color appearance and potential motion discontinuity. To recover the realistic color appearance, we force the adversarial loss \cite{zhu2017unpaired} to make the fused images look natural:
\begin{equation}
	\setlength\abovedisplayskip{0pt}
  \setlength\belowdisplayskip{0pt}
\begin{aligned}
\mathcal{L}_{adv} = \mathbb{E}[{log(1-D(\textbf{\emph{${\tilde{I}_{t}}$}}))}] + \mathbb{E}[{log(1-D(\textbf{\emph{${\tilde{I}_{t + \Delta t}}$}}))}],
 \label{eq:Adversarial}
 \end{aligned}
\end{equation}
where $D$ is the discriminator. To prevent the motion discontinuity, we explore a motion-related prior knowledge to constrain the fusion process. We start ideally to yield the optical flow basic model as follows via Taylor expansion:
\begin{equation}\footnotesize
  \setlength\abovedisplayskip{0pt}
  \setlength\belowdisplayskip{0pt}
\begin{aligned}
I(x+dx, t+dt) = I(x, t) + (\frac{\partial{I}}{\partial{x}}dx + \frac{\partial{I}}{\partial{t}}dt) + O(dx, dt).
  \label{eq:taylor_expansion}
\end{aligned}
\end{equation}

Then, we remove the high-order error term ${O(dx, dt)}$ to approximate Eq. \ref{eq:taylor_expansion} as follows:
\begin{equation}
  \setlength\abovedisplayskip{1pt}
  \setlength\belowdisplayskip{1pt}
\begin{aligned}
I_t = - \nabla{I} \cdot U,
  \label{eq:constancy_equalization}
\end{aligned}
\end{equation}
where ${U}$ denotes optical flow, ${- \nabla{I} \cdot U}$ is the spatiotemporal gradient of current frame. ${I_t}$ denotes temporal brightness change, which can be approximated as the accumulated events warped by optical flow in a certain time, namely ${I_t = U\cdot\sum\nolimits_{e_i \in \Delta{t}}p_{i}C}$. Therefore, we take the spatiotemporal gradient consistency to constrain the fusion:
\begin{equation}\footnotesize
  \setlength\abovedisplayskip{1pt}
  \setlength\belowdisplayskip{1pt}
\begin{aligned}
\mathcal{L}_{consis} = \sum\nolimits ||U\cdot\sum\nolimits_{e_i \in \Delta{t}}p_{i}C + \nabla{I} \cdot U||_1 \odot V / \sum\nolimits V,
 \label{eq:self_flow}
 \end{aligned}
\end{equation}
where $V$ is the valid mask generated by optical flow $U$. The visual luminance fusion that obeys the basic motion model can guarantee the valid motion feature matching.

\begin{figure}
  \setlength{\abovecaptionskip}{5pt}
  \setlength{\belowcaptionskip}{-5pt}
  \centering
   \includegraphics[width=0.99\linewidth]{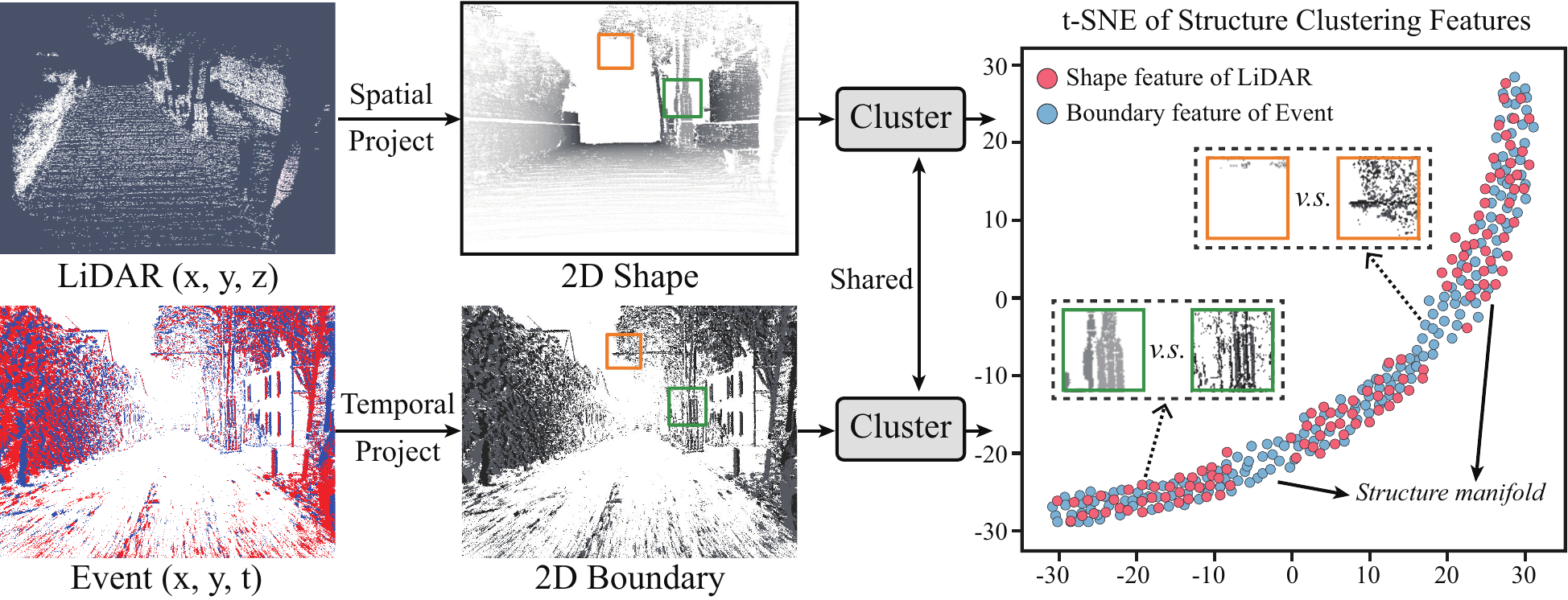}
   \caption{Clustering feature distribution of event and LiDAR. Event and LiDAR share the same structure manifold, where the boundary distribution of event is continuous while the shape distribution of LiDAR is truncated. This motivates us to take the structure as a homogeneous space to fuse the boundary-shape knowledge.
   }
   \label{Cluster}

\end{figure}

\subsection{Visual Structure Fusion}
LiDAR point cloud has the incomplete contour of the scene due to a non-uniform laser beam, limiting the scene flow performance. We aim to introduce the event to assist in completing the physical structure of the LiDAR point cloud.

\noindent
\textbf{Local Boundary \emph{v.s.} Global Shape.}
LiDAR emits uniform laser beams in the horizontal direction, and the reflected laser can show the global shape of ${360^\circ}$ scene. However, LiDAR just emits the non-uniform laser beams in the vertical direction due to the vertical angle. The larger the vertical angle, the fewer the number of the reflected point cloud, resulting in the incomplete contour of LiDAR. In contrast, event can sense the rich motion boundaries of the scene. Therefore, the local boundary of event and the global shape of LiDAR can jointly complete the intrinsic structure of the scene. To illustrate the complementary nature between LiDAR and event in the structure space, as shown in Fig. \ref{Cluster}, we cluster the spatially-projected 2D LiDAR frame and the temporally-projected 2D event frame into the same structure manifold, and visualize the clustering feature distributions of LiDAR and event. First, the distribution trend of LiDAR is similar to the event. Second, the distribution of event boundary is continuous while the distribution of LiDAR shape is truncated, indicating that the contours of the 2D LiDAR frame are incomplete but those of the 2D event frame are relatively complete. This motivates us to take the scene structure as a homogeneous space to fuse the local boundary of event into the global shape of LiDAR for physical structure integrity.

\noindent
\textbf{Self-Similarity Clustering.}
Structure fusion mainly depends on a self-similarity clustering strategy, which is divided into two steps: intra-modal structure neighbor and inter-modal data fusion. The former is to properly cluster LiDAR shape points and event boundary points into the same neighbor. The latter is to map the local boundary coordinates of event to the global shape coordinates of LiDAR in the neighbor, thus completing the physical contour. Given the aligned LiDAR point cloud $pc=\{x_i, y_i, z_i | i \in M\}$, the corresponding event stream $ev=\{x_j, y_j, t_j, p_j | j \in N\}$, intrinsic parameter $K=\{f, c_x, c_y\}$, we first normalize the event stream along the timestamps to obtain the 2D coordinates of event $P_e=\{u_j=x_j, v_j=y_j, p_j | \{x_j, y_j, p_j\} \in ev\}$, and use the intrinsic parameter to project the LiDAR point cloud into the event camera coordinate system $P_l = \{u_i, v_i, d_i\}$:
\begin{equation}
  \setlength\abovedisplayskip{1pt}
  \setlength\belowdisplayskip{1pt}
\begin{aligned}
 &u_i = f \cdot {x_i}/{z_i} + c_x, v_i = f \cdot {y_i}/{z_i} + c_y, d_i = z_i,\\
s&.t. \{x_i, y_i, z_i\} \in pc, 0 < u_i < w, 0 < u_i < h,
 \label{eq:project}
 \end{aligned}
\end{equation}
where $w$, $h$ is the size of event frame, $f$ is focal length, $c_x$,$c_y$ are x, y-axis offsets. In the intra-modal structure neighbor stage, we construct the distance function as the self-similarity measurement metric for 2D coordinates of the event:
\begin{equation}\footnotesize
  \setlength\abovedisplayskip{1pt}
  \setlength\belowdisplayskip{1pt}
\begin{aligned}
 &d_{p} = \sqrt{(p_1 - p_2)^2}, d_{s} = \sqrt{(u_1 - u_2)^2+(v_1-v_2)^2}, \\
 &D_{e} = \sqrt{d^2_p + ({d_{s}}/{N_s})^2}, s.t. \{u_{1,2}, v_{1,2}, p_{1,2}\} \in P_e,
 \label{eq:distance}
 \end{aligned}
\end{equation}
where $N_s$ is the max distance of 2D coordinates. We also define the same distance function $D_{l}$ for the 2D coordinates of LiDAR. We iteratively optimize the distance functions $D_e$, $D_l$ to obtain the neighbors $U$, which represent the region-level structure information like the image superpixel. In the inter-modal data fusion stage, we propose a similarity-based bi-direction fusion strategy: event$\rightarrow$LiDAR structure fusion and LiDAR$\rightarrow$event depth fusion. As for the event$\rightarrow$LiDAR structure fusion, we take the various 2D LiDAR coordinates of each neighbor as the center, and use the K-nearest neighbor to select the top $k$ 2D event coordinates with the highest similarity and fill them into the 2D LiDAR coordinates in the same neighbor. As for the LiDAR$\rightarrow$event depth fusion, we filter the top $k$ 2D LiDAR coordinates in the same way, and weightedly fuse their depths to the selected event coordinates to obtain the final depth map. To ensure the back-propagation of the whole framework, we introduce a cross-attention transformer \cite{vaswani2017attention} to model the visual structure fusion process between event and LiDAR. We use the weighted-fused depth maps as the pseudo labels ${d}^{pse}_t$, ${d}^{pse}_{t+\Delta t}$ to train the cross-attention transformer:
\begin{equation}\footnotesize
  \setlength\abovedisplayskip{1pt}
  \setlength\belowdisplayskip{1pt}
\begin{aligned}
\mathcal{L}_{pse} = \sum\nolimits ||\tilde{d}_t - {d}^{pse}_t||_1 + ||\tilde{d}_{t+\Delta t} - {d}^{pse}_{t+\Delta t}||_1,
 \label{eq:pseudo_loss}
 \end{aligned}
\end{equation}
where $\tilde{d}_t$, $\tilde{d}_{t+\Delta t}$ are the fused adjacent depth maps estimated from the transformer network, and then are inversely transformed to the 3D point clouds $\tilde{pc}_t$, $\tilde{pc}_{t+\Delta t}$.

\begin{figure}
  \setlength{\abovecaptionskip}{5pt}
  \setlength{\belowcaptionskip}{-5pt}
  \centering
   \includegraphics[width=0.99\linewidth]{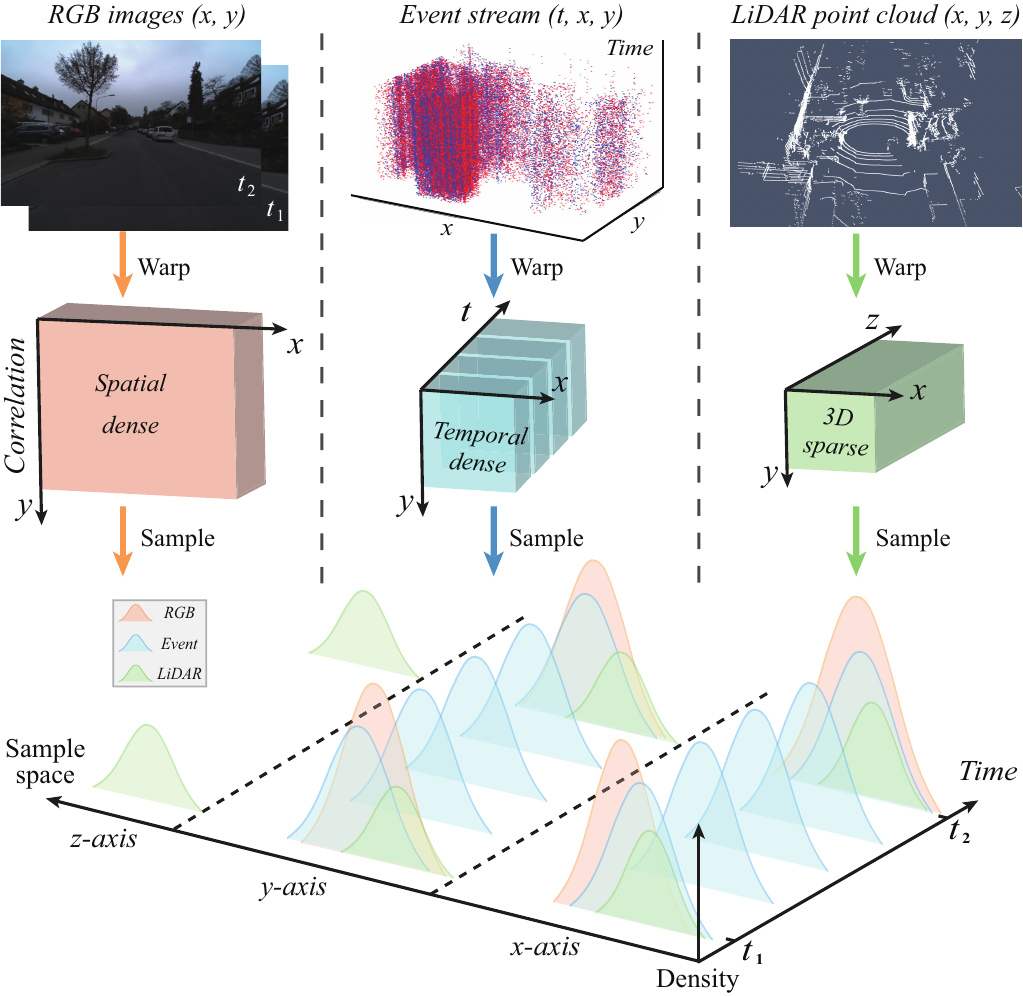}
   \caption{Correlation distributions of RGB, event and LiDAR. The three modalities have similar distributions in x, y-axis correlation, with z-axis correlation unique to LiDAR. RGB correlation is spatially dense, event is temporally dense, while LiDAR is spatiotemporally sparse. This inspires us to build the homogeneous correlation space to fuse the complementary motion knowledge.
   }
   \label{Motion_Distribution}

\end{figure}
\subsection{Motion Correlation Fusion}
Visual luminance and structure fusions do ensure the visualization of RGB and LiDAR, while how to physically and complementarily fuse the mapped motion features of various modalities to improve scene flow is a challenge.

\noindent
\textbf{Spatiotemporal Dense \emph{v.s.} Sparse Correlation.}
Motion space that scene flow relies on consists of two types: temporal correlation subspace \cite{teed2020raft} and decoded motion subspace \cite{jiang2021transformer}. The former is to model the feature similarity between adjacent frames, and the latter contains the features decoded by recurrent network (\emph{e.g.}, GRU \cite{cho2014learning}). Since the correlation subspace has more physical meaning, it has the potential to be a homogeneous space. As far as we know, RGB provides texture information with high spatial resolution, event provides boundary information with high temporal resolution, and LiDAR provides 3D (x, y, z) scene structure information. This inspires us to consider that, in the correlation space, RGB may promise the x, y-axis spatial-dense correlation, event may ensure the x, y-axis temporal-dense correlation and LiDAR may guarantee the x, y, z-axis accurate but sparse correlation. To verify our insight, as shown in Fig. \ref{Motion_Distribution}, we transform the visual RGB, event and LiDAR features into the correlation subspace of motion space via warp operator \cite{sun2018pwc}, and calculate their distributions in x, y, z three dimensions, respectively. Note that, the correlation distribution of event is analyzed along the timestamps. On one hand, the correlation distribution trends of these three modalities are quite similar along the x and y axes, with z-axis correlation unique to LiDAR. On the other hand, in terms of the complementarity in the correlation space, RGB is spatial-dense, event is temporal-dense, and LiDAR is spatiotemporal-sparse. Therefore, we treat the correlation space as a homogeneous space to fuse the cross-modal knowledge for motion continuity.

\begin{table*}\footnotesize
    \setlength{\abovecaptionskip}{4pt}
    \setlength\tabcolsep{2pt}
    \setlength{\belowcaptionskip}{-7pt}

  \centering
  \renewcommand\arraystretch{1.0}
  \begin{tabular}{cccccccccccc}
  \Xhline{1pt}
  \multicolumn{2}{c|}{\multirow{2}{*}{Method}}&
  \multicolumn{6}{c|}{\multirow{1}{*}{Scene flow methods}}&
  \multicolumn{4}{c}{\multirow{1}{*}{Optical flow methods}} \\

  \cline{3-12}
  \multicolumn{2}{c|}{}& \multicolumn{1}{c|}{\multirow{1}{*}{RAFT-3D}} & \multicolumn{1}{c|}{\multirow{1}{*}{RAFT-3D w/ e}} & \multicolumn{1}{c|}{\multirow{1}{*}{PV-RAFT}} & \multicolumn{1}{c|}{\multirow{1}{*}{CamLiFlow}} & \multicolumn{1}{c|}{\multirow{1}{*}{RPEFlow}} & \multicolumn{1}{c|}{\multirow{1}{*}{\textbf{VisMoFlow}}} &
  \multicolumn{1}{c|}{\multirow{1}{*}{SMURF}}&
  \multicolumn{1}{c|}{\multirow{1}{*}{FlowFormer}}&
  \multicolumn{1}{c|}{\multirow{1}{*}{RPEFlow}}&
  \multicolumn{1}{c}{\multirow{1}{*}{\textbf{VisMoFlow}}} \\

  \hline
  \multicolumn{2}{c|}{\multirow{1}{*}{Input}}&  \multicolumn{1}{c|}{RGB} & \multicolumn{1}{c|}{RGB} & \multicolumn{1}{c|}{PC}& \multicolumn{1}{c|}{RGB+PC}&  \multicolumn{1}{c|}{RGB+PC+EV}& \multicolumn{1}{c|}{\textbf{RGB+PC+EV}} & \multicolumn{1}{c|}{RGB} & \multicolumn{1}{c|}{RGB} & \multicolumn{1}{c|}{RGB+PC+EV} & \multicolumn{1}{c}{\textbf{RGB+PC+EV}}\\

  \hline
   \multicolumn{1}{c|}{\multirow{2}{*}{Day}}& \multicolumn{1}{c|}{EPE} & \multicolumn{1}{c|}{0.095} & \multicolumn{1}{c|}{--} & \multicolumn{1}{c|}{0.055}& \multicolumn{1}{c|}{0.033}&  \multicolumn{1}{c|}{0.060}& \multicolumn{1}{c|}{\textbf{0.012}} & \multicolumn{1}{c|}{2.01} & \multicolumn{1}{c|}{0.607} & \multicolumn{1}{c|}{0.556} & \multicolumn{1}{c}{\textbf{0.198}}\\

  \cline{2-12}
  \multicolumn{1}{c|}{} & \multicolumn{1}{c|}{ACC} & \multicolumn{1}{c|}{73.48\%} & \multicolumn{1}{c|}{--} & \multicolumn{1}{c|}{79.96\%}& \multicolumn{1}{c|}{91.40\%}&  \multicolumn{1}{c|}{81.73\%}& \multicolumn{1}{c|}{\textbf{98.55\%}} & \multicolumn{1}{c|}{83.34\%} & \multicolumn{1}{c|}{89.08\%} & \multicolumn{1}{c|}{88.98\%} & \multicolumn{1}{c}{\textbf{97.11\%}}\\

  \hline
  \multicolumn{1}{c|}{\multirow{2}{*}{Night}} & \multicolumn{1}{c|}{EPE} & \multicolumn{1}{c|}{0.112} & \multicolumn{1}{c|}{0.104} & \multicolumn{1}{c|}{0.055}& \multicolumn{1}{c|}{0.047}&  \multicolumn{1}{c|}{0.056}& \multicolumn{1}{c|}{\textbf{0.027}} & \multicolumn{1}{c|}{11.360} & \multicolumn{1}{c|}{2.085} & \multicolumn{1}{c|}{0.716} & \multicolumn{1}{c}{\textbf{0.353}}\\

  \cline{2-12}
  \multicolumn{1}{c|}{}& \multicolumn{1}{c|}{ACC} & \multicolumn{1}{c|}{65.65\%} & \multicolumn{1}{c|}{72.65\%} & \multicolumn{1}{c|}{79.97\%}& \multicolumn{1}{c|}{85.62\%}&  \multicolumn{1}{c|}{81.47\%}& \multicolumn{1}{c|}{\textbf{95.62\%}} & \multicolumn{1}{c|}{55.12\%} & \multicolumn{1}{c|}{77.05\%} & \multicolumn{1}{c|}{78.85\%} & \multicolumn{1}{c}{\textbf{96.28\%}}\\

  \Xhline{1pt}
  \end{tabular}
  \caption{Quantitative results on synthetic Event-KITTI dataset. ``PC'' is LiDAR point cloud, ``EV'' is event. ``w/ e'' is image enhancement.}
   \label{tab:quantitative_synthetic_result}
\end{table*}

\begin{figure*}
  \setlength{\abovecaptionskip}{2pt}
  \setlength{\belowcaptionskip}{-10pt}
  \centering
   \includegraphics[width=0.99\linewidth]{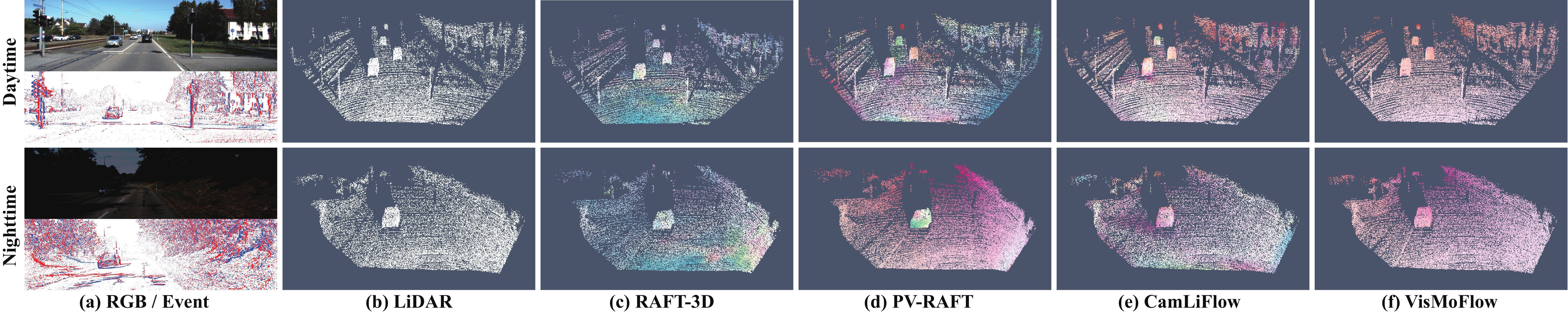}
   \caption{Visual comparison of scene flows on synthetic Event-KITTI dataset.
   }
   \label{Comparison}

\end{figure*}

\noindent
\textbf{Motion Distribution Alignment.}
Correlation fusion is first to fuse the dense-sparse complementary knowledge between various modalities in the single-axis correlation, and then merge the fused correlation distributions across all three axes. First, we voxelize the event stream into multiple-slice voxels. We encode the fused RGB images, the voxelized events and the fused LiDAR point clouds to the RGB correlation $cv_r$, the event correlation $cv_e$ and the LiDAR correlation $cv_l$ using the warp operator. Then, we randomly sample $N$ 3D points from LiDAR. From these points, we obtain 2D coordinates for RGB and event with intrinsic parameters. Next, we use the sampled coordinates as the center, and perform 2D spatially-sampling for the RGB correlation to obtain the corresponding spatial-dense x, y-axis correlations $cv^x_r$, $cv^y_r$. As for the event, we temporally sample the correlation features to get the temporal-dense x, y-axis correlations $cv^x_e$, $cv^y_e$. For the LiDAR, we spatially sample the correlations into the relatively sparse x, y, z-axis correlations $cv^x_l$, $cv^y_l$, $cv^z_l$. Before the correlation fusion, we apply K-L divergence to align the multimodal motion correlation distributions:
\begin{equation}\scriptsize
  \setlength\abovedisplayskip{1pt}
  \setlength\belowdisplayskip{1pt}
\begin{aligned}
\mathcal{L}^{kl}_{corr} = \sum\nolimits \Phi(cv^{x,y}_l) \cdot log\frac{\Phi(cv^{x,y}_l)}{\Phi(cv^{x,y}_r)} + \Phi(cv^{x,y}_l) \cdot log\frac{\Phi(cv^{x,y}_l)}{\Phi(cv^{x,y}_e)},
  \label{eq:kl_cost_x}
\end{aligned}
\end{equation}
where ${\Phi}$ is the softmax function. We first fuse the aligned x, y-axis correlations between various modalities and then concatenate the z-axis correlations of LiDAR:
\begin{equation}\footnotesize
  \setlength\abovedisplayskip{1pt}
  \setlength\belowdisplayskip{1pt}
\begin{aligned}
corr = Concat\{\frac{1}{T}\sum\nolimits^T_{i=0}(cv^x_r+cv^{x,i}_e+cv^x_l)/3, \\
\frac{1}{T}\sum\nolimits^T_{i=0}(cv^y_r+cv^{y,i}_e+cv^y_l)/3, cv^z_l\},
  \label{eq:kl_cost}
\end{aligned}
\end{equation}
where $T$ denotes the number of the temporal slices of events. Furthermore, we introduce GRU to recursively decode the fused correlation features to estimate the final scene flow. In addition, we learn the scene flow and the corresponding optical flow with photometric loss \cite{jason2016back}:
\begin{equation}\footnotesize
  \setlength\abovedisplayskip{1pt}
  \setlength\belowdisplayskip{1pt}
\begin{aligned}
\mathcal{L}_{pho} &= \sum\nolimits{\psi(\tilde{I}_{t} - w_{2D}(\tilde{I}_{t+\Delta t}))}\odot V_{2D}/\sum\nolimits{V_{2D}} \\
&+ \sum\nolimits{\psi(\tilde{pc}_{t} - w_{3D}(\tilde{pc}_{t+\Delta t}))}\odot V_{3D}/\sum\nolimits{V_{3D}},
  \label{eq:photo_loss}
\end{aligned}
\end{equation}
where $\psi$ is a $L_p$ norm ($p = 0.4$). $V_{2D}$, $V_{3D}$ is the 2D and 3D occlusion masks by checking forward-backward consistency \citep{zou2018df}. We also apply the photometric loss to train the event motion feature encoder. The proposed motion correlation fusion can ensure 3D motion spatiotemporal continuity.

\begin{table}\footnotesize
    \setlength{\abovecaptionskip}{4pt}
    \setlength\tabcolsep{0pt}
    \setlength{\belowcaptionskip}{-5pt}

  \centering
  \renewcommand\arraystretch{1.1}
  \begin{tabular}{ccccccc}
  \Xhline{1pt}
  \multicolumn{2}{c|}{\multirow{1}{*}{Method}}&
  \multicolumn{1}{c|}{\multirow{1}{*}{RAFT-3D}} &
   \multicolumn{1}{c|}{\multirow{1}{*}{PV-RAFT}} & \multicolumn{1}{c|}{\multirow{1}{*}{CamLiFlow}} & \multicolumn{1}{c|}{\multirow{1}{*}{RPEFlow}} & \multicolumn{1}{c}{\multirow{1}{*}{\textbf{VisMoFlow}}} \\

  \hline
  \multicolumn{2}{c|}{\multirow{1}{*}{Input}}&  \multicolumn{1}{c|}{RGB}& \multicolumn{1}{c|}{PC}& \multicolumn{1}{c|}{RGB+PC}&  \multicolumn{1}{c|}{RGB+PC+EV}& \multicolumn{1}{c}{\textbf{RGB+PC+EV}}\\

  \hline
  \multicolumn{1}{c|}{\multirow{2}{*}{Day}}& \multicolumn{1}{c|}{EPE} & \multicolumn{1}{c|}{0.167}& \multicolumn{1}{c|}{0.183}&  \multicolumn{1}{c|}{0.113}& \multicolumn{1}{c|}{0.103}& \multicolumn{1}{c}{\textbf{0.084}}\\

  \cline{2-7}
  \multicolumn{1}{c|}{} & \multicolumn{1}{c|}{ACC}  & \multicolumn{1}{c|}{13.16\%}& \multicolumn{1}{c|}{37.28\%}&  \multicolumn{1}{c|}{55.69\%}& \multicolumn{1}{c|}{60.81\%}& \multicolumn{1}{c}{\textbf{70.34\%}} \\

  \hline
  \multicolumn{1}{c|}{\multirow{2}{*}{Night}}& \multicolumn{1}{c|}{EPE} &  \multicolumn{1}{c|}{0.359}& \multicolumn{1}{c|}{0.190}& \multicolumn{1}{c|}{0.125}&  \multicolumn{1}{c|}{0.094}& \multicolumn{1}{c}{\textbf{0.090}}\\

  \cline{2-7}
  \multicolumn{1}{c|}{} & \multicolumn{1}{c|}{ACC} & \multicolumn{1}{c|}{5.04\%}& \multicolumn{1}{c|}{40.98\%}& \multicolumn{1}{c|}{53.10\%}&  \multicolumn{1}{c|}{66.49\%}& \multicolumn{1}{c}{\textbf{68.31\%}} \\

  \Xhline{1pt}
  \end{tabular}
  \caption{Quantitative results on real DSEC dataset.}
   \label{tab:quantitative_real_result}
\end{table}

\subsection{Optimization and Implementation Details}
Consequently, the total objective for the proposed hierarchical fusion framework is written as follows:
\begin{equation}\footnotesize
  \setlength\abovedisplayskip{1pt}
  \setlength\belowdisplayskip{1pt}
\begin{aligned}
\mathcal{L} = \mathcal{L}_{pho} + \lambda_1\mathcal{L}_{adv} + \lambda_2\mathcal{L}_{consis} + \lambda_3\mathcal{L}_{pse} + \lambda_4\mathcal{L}^{kl}_{corr},
  \label{eq:total_loss}
\end{aligned}
\end{equation}
where [$\lambda_1, ..., \lambda_4$] are the weights that control the importance of the related losses. The first term is to maintain the capability of motion feature encoders of RGB, event and LiDAR. The second and third terms are to constrain the Event-RGB visual luminance fusion, the fourth term is to promise that the cross-attention transformer can learn to fuse the Event-LiDAR structure, and the intention of the last term aims to promote the RGB-Event-LiDAR motion correlation fusion. We set the KNN selection number $k$ as 5, the event temporal slices $T$ as 10, and the sample number $N$ as 1000. During the training phase, we only need three steps. First, we train Event-RGB visual luminance fusion for high dynamic imaging. Second, we train Event-LiDAR visual structure fusion for physical structure integrity. Finally, we train RGB-Event-LiDAR motion correlation fusion for 3D motion spatiotemporal continuity. Note that, the proposed method is first pre-trained to initialize the motion estimators in a supervised manner. During the testing phase, the whole framework consists of the luminance fusion network, the cross-attention transformer and motion estimation networks, thus achieving the end-to-end estimation of scene flow.

\begin{figure*}
  \setlength{\abovecaptionskip}{2pt}
  \setlength{\belowcaptionskip}{-10pt}
  \centering
   \includegraphics[width=0.99\linewidth]{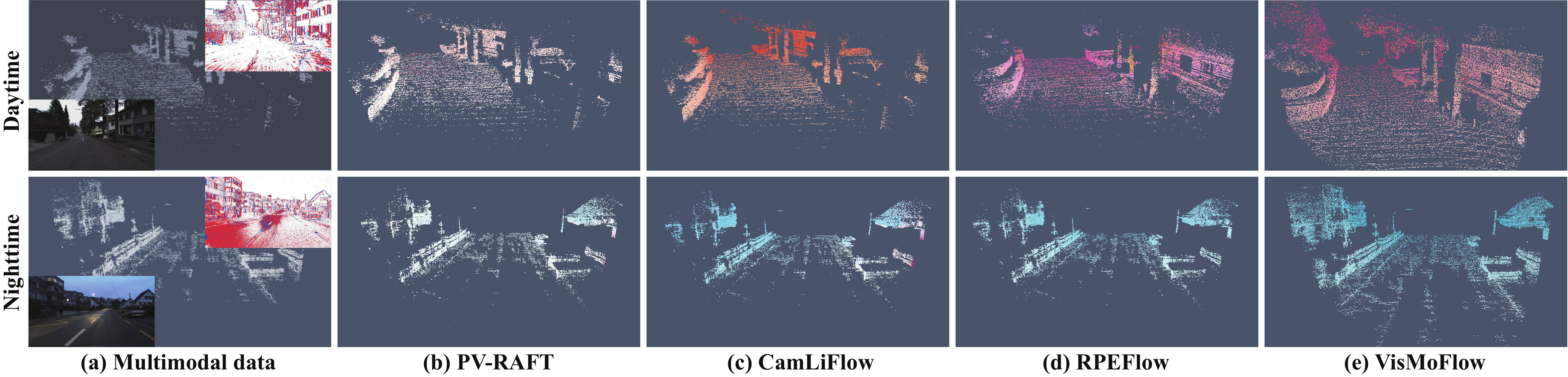}
   \caption{Visual comparison of scene flows on real DSEC dataset.
   }
   \label{Real_Comparison}

\end{figure*}

\section{Experiments}
\label{sec:experiment}
\subsection{Experiment Setup}
\noindent
\textbf{Dataset.}
We conduct extensive experiments on one synthetic (\emph{e.g.}, Event-KITTI) and one real (\emph{e.g.}, DSEC) datasets, including daytime and nighttime scenes.

\noindent
$\bullet$ \textbf{Event-KITTI.}
The origin KITTI \cite{menze2015object} is set in daytime scenes. We take the V2E model \cite{hu2021v2e} to generate the event stream from daytime images, and use the noise model \cite{zheng2020optical} to synthesize nighttime images corresponding to daytime.

\noindent
$\bullet$ \textbf{DSEC.}
DSEC is a multimodal dataset \cite{Gehrig21ral}, which covers the multimodal data of RGB camera, LiDAR and event camera in various outdoor scenes and various times.

\noindent
\textbf{Comparison Methods.}
We choose optical flow and scene flow methods for comparison in both daytime and nighttime scenes. As for optical flow comparison, we choose unsupervised (SMURF \cite{stone2021smurf}), supervised (FlowFormer \cite{huang2022flowformer}), and multimodal fusion (RPEFlow \cite{wan2023rpeflow}) methods. As for scene flow comparison, we choose RGB-based unimodal (RAFT-3D \cite{teed2021raft}), LiDAR-based unimodal (PV-RAFT \cite{wei2021pv}), RGB-LiDAR multimodal fusion (CamLiFlow \cite{liu2022camliflow}) and RGB-Event-LiDAR multimodal fusion (RPEFlow) methods. Note that, in nighttime scenes, we add an additional comparison: first performing image enhancement (\emph{e.g.}, KinD++ \cite{zhang2019kindling}) and then estimate motion on the enhanced results (named as ``w/ e'').  We choose the average end-point error (EPE \cite{menze2015object}) and the percentage of flow accuracy within 1px for optical flow or 5cm for scene flow (ACC \cite{liu2022camliflow}) as the evaluation metrics.

\begin{table}
    \setlength{\abovecaptionskip}{5pt}
    \setlength\tabcolsep{7pt}
    \setlength{\belowcaptionskip}{-5pt}
  \centering
  \renewcommand\arraystretch{1.0}
  \begin{tabular}{c|cc}
    \Xhline{1pt}
      Fusion strategy & EPE & ACC \\
      \hline
      w/o any fusion & 0.151 & 49.20\% \\
       w/ visual fusion & 0.118 & 54.35\%\\
       w/ motion fusion & 0.098 & 62.87\%\\
      w/ visual-motion fusion & \textbf{0.084} & \textbf{70.34\%}\\
       \Xhline{1pt}
  \end{tabular}
  \caption{Effectiveness of hierarchical visual-motion fusion.}
   \label{tab:fusion_framework}
\end{table}

\subsection{Comparison Experiment}
\noindent
\textbf{Comparison on Synthetic Dataset.}
In Table \ref{tab:quantitative_synthetic_result} and Fig. \ref{Comparison}, we compare different motion methods on the synthetic Event-KITTI dataset. First, the proposed method significantly outperforms the competing methods in optical flow and scene flow. Second, in nighttime scenes, the performance of those methods that use RGB images decreases slightly. Note that, image enhancement can indeed promote the RGB-based motion methods in nighttime scenes, but the improvement is limited. Third, multimodal methods are more effective than unimodal methods in improving motion estimation due to the complementary knowledge between various modalities. Notably, the proposed visual-motion fusion surpasses the RPEFlow that relies solely on motion fusion.

\noindent
\textbf{Comparison on Real Scenes.}
In Table \ref{tab:quantitative_real_result}, we compare the scene flow methods in real daytime and nighttime scenes on the DSEC dataset. We have two conclusions. First, multimodal methods are far superior to unimodal methods. This is because the complementary knowledge of various modalities in real scenes can make up for their own disadvantage, conducive to deeply exploiting the characteristic of scene flow. Second, event camera can further effectively improve nighttime scene flow. The main reason is that, event camera have the advantage of high dynamic range to sense scene motion at night. In Fig. \ref{Real_Comparison}, our scene flow visualization results show more complete structures and perform better than other competing methods. The main reason is that, the visual structure fusion does fuse the local boundary of event into the global shape of LiDAR for point cloud physical integrity, thus improving the matching capability of motion features.

\begin{figure}
  \setlength{\abovecaptionskip}{2pt}
  \setlength{\belowcaptionskip}{-8pt}
  \centering
   \includegraphics[width=0.99\linewidth]{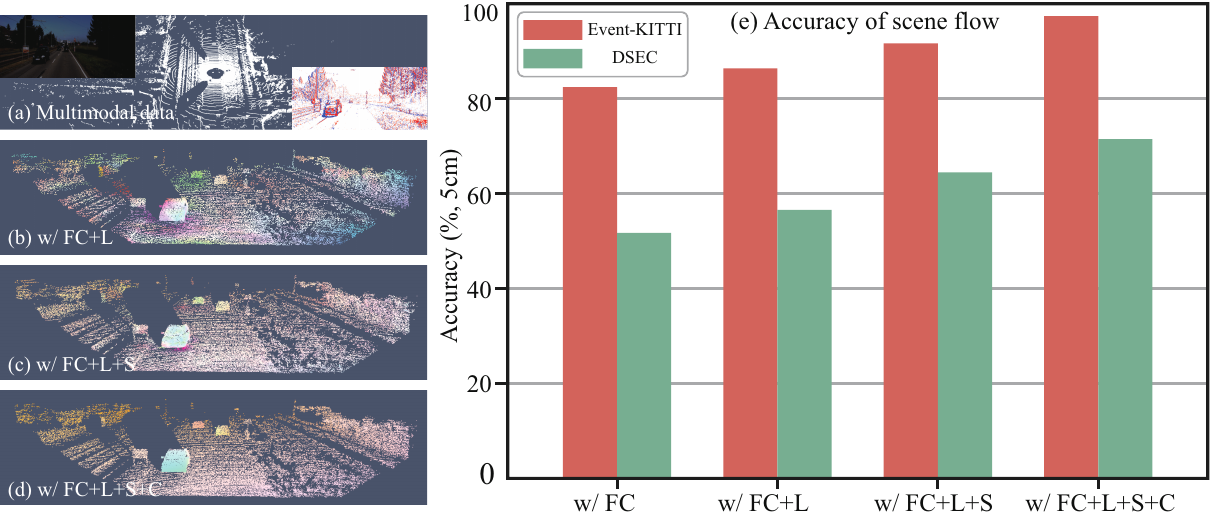}
   \caption{Effectiveness of homogeneous spaces in fusion. (a) Input data. (b-d) Scene flow. ``FC'' is implicit feature concatenate. ``L'' is luminance space. ``S'' is structure space. ``C'' is correlation space.
   }
   \label{homogeneousSpace}

\end{figure}

\subsection{Ablation Study}
\noindent
\textbf{How does Hierarchical Visual-Motion Fusion Work?}
In Table \ref{tab:fusion_framework}, we demonstrate the effectiveness of the proposed hierarchical visual-motion fusion framework. With only visual fusion, there is an improvement in scene flow. With only motion fusion, scene flow is significantly improved. When we apply visual fusion and motion fusion, the hierarchical fusion is superior to the motion fusion alone. Therefore, motion fusion is the key to promoting scene flow, and visual fusion can update the upper limit of scene flow performance.

\noindent
\textbf{The Role of homogeneous Space in Fusion.}
We illustrate the impact of different homogeneous spaces on scene flow in Fig. \ref{homogeneousSpace}. When only implicit feature fusion (\emph{i.e.}, feature concatenate (FC)), more errors of the scene flow are caused by feature mismatching. With visual luminance space (w/ FC+L), scene flow is improved due to the enhanced visual RGB features. Visual structure space (w/ FC+L+S) further promotes the integrity of the scene flow structure. With motion correlation space (w/ FC+L+S+C), the motion metric is greater, and the scene flow visualization is spatiotemporally smoother. Therefore, the homogeneous space can serve as a guide to physically facilitate the fusion process.

\begin{table}
    \setlength{\abovecaptionskip}{4pt}
    \setlength\tabcolsep{7pt}
    \setlength{\belowcaptionskip}{-8pt}
  \centering
  \renewcommand\arraystretch{1.05}
  \begin{tabular}{ccc|cc}
    \Xhline{1pt}
      $\mathcal{L}_{consis}$ & $\mathcal{L}_{pse}$ & $\mathcal{L}^{kl}_{corr}$& EPE & ACC \\
			 \hline
             $\times$& $\times$&$\times$& 0.122 &53.17\% \\
		 $\surd$& $\times$&$\times$&  0.112 & 56.41\% \\

		$\times$ 	&$\surd$ &$\times$& 0.107 & 58.25\% \\

		$\times$ &$\times$ &$\surd$ & 0.092 & 65.43\% \\
		$\surd$ 	&$\surd$ &$\surd$ & \textbf{0.084} &\textbf{70.34\%} \\
       \Xhline{1pt}
  \end{tabular}
  \caption{Ablation study on fusion losses.}
   \label{tab:loss}
\end{table}

\begin{figure}
  \setlength{\abovecaptionskip}{2pt}
  \setlength{\belowcaptionskip}{-10pt}
  \centering
   \includegraphics[width=0.99\linewidth]{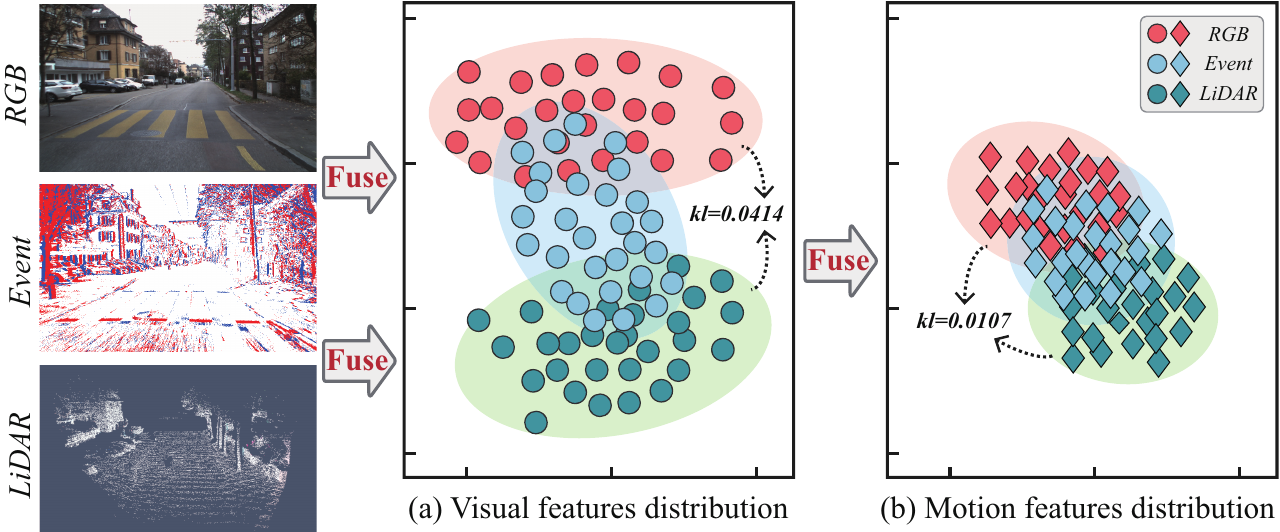}
   \caption{t-SNE visualization of visual and motion features. In (a) visual space, there exists a gap between RGB and LiDAR, while event serves as a bridge across the two modalities. In (b) motion space, the three modalities are further close to each other.
   }
   \label{Bridge}

\end{figure}

\noindent
\textbf{Effectiveness of Fusion Losses.}
In Table \ref{tab:loss}, we conduct ablation studies on the main fusion losses. Without any fusion losses, the whole framework still has the visual-motions fusion process. Each loss improves scene flow to a certain extent, while the motion distribution alignment loss $\mathcal{L}^{kl}_{corr}$ plays a key role in scene flow. The three losses can jointly promote the visual$\rightarrow$motion progressive fusion process.

\subsection{Discussion}
\noindent
\textbf{How does Event Play the Bridge.}
In Fig. \ref{Bridge}, we illustrate the importance of event as a bridge between RGB and LiDAR in visual and motion feature spaces via t-SNE. In the visual space, there exists a large gap between RGB and LiDAR features, while event features can just serve as a bridge across the two modalities. When the fused visual features are mapped to the motion space, the motion features of RGB and LiDAR are close to each other and have a certain intersection, which contains most of the event motion features. Therefore, the event can reasonably bridge the RGB and LiDAR.

\noindent
\textbf{Effectiveness of Spatiotemporal Gradient Consistency.}
We analyze the impact of spatiotemporal gradient consistency on the Event-RGB fused image result and the optical flow result from the training process in Fig. \ref{gradient_consistency}. Without the consistency loss, the training curve fluctuates violently, the fused RGB image looks clear while there are some artifacts in the optical flow. With the consistency loss, the fluctuation of the training curve is smoother, the fused RGB image is also clear and the optical flow boundaries become sharp. This shows that spatiotemporal gradient consistency does constrain the mapping relationship between the Event-RGB fused visual features and the motion features.

\noindent
\textbf{Why Structure Neighbor is Necessary?}
Structure neighbor is to extract the region-level structure from the pixel-level structure, which has the local self-similarity characteristics. In Table \ref{tab:neighbor}, without any neighbor, the scene flow is slightly limited. This is because the full pixel association between event and LiDAR increases the probability of local-global structure feature mismatching. With grid-based neighbor, the scene flow performance is improved but is lower than the clustering-based neighbor. The main reason is that, the grid-based neighbor may contain partial information of different objects, leading to the structure fusion bias, while the clustering-based neighbor can reduce this error.

\begin{table}
    \setlength{\abovecaptionskip}{4pt}
    \setlength\tabcolsep{5pt}
    \setlength{\belowcaptionskip}{-8pt}
  \centering
  \renewcommand\arraystretch{1.05}
  \begin{tabular}{c|cc}
    \Xhline{1pt}
      Structure neighbor strategy & EPE & ACC \\
      \hline
      w/o any neighbor & 0.095 & 64.70\% \\
       w/ grid-based neighbor & 0.088 & 69.05\%\\
      w/ clustering-based neighbor & \textbf{0.084} & \textbf{70.34\%}\\
       \Xhline{1pt}
  \end{tabular}
  \caption{Choice of different structure neighbor strategies.}
   \label{tab:neighbor}
\end{table}

\begin{figure}
  \setlength{\abovecaptionskip}{2pt}
  \setlength{\belowcaptionskip}{-5pt}
  \centering
   \includegraphics[width=0.99\linewidth]{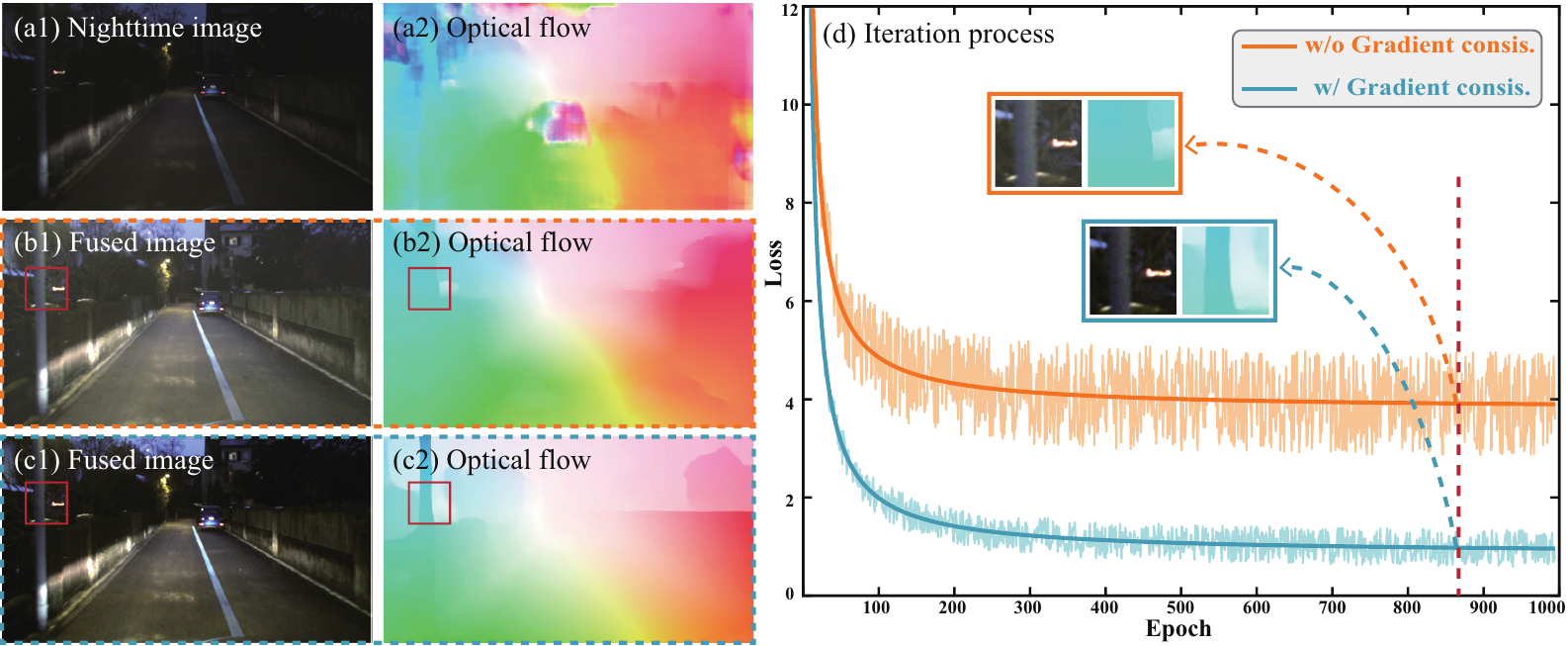}
   \caption{Effect of spatiotemporal gradient consistency on the fused image and optical flow. (a) Without fusion (b) Fusion without gradient consistency. (c) Fusion with gradient consistency. The gradient consistency can smooth training curve, conducive to constraining the mapping relationship between the fused image and optical flow.
   }
   \label{gradient_consistency}

\end{figure}

\noindent
\textbf{Limitation.}
The proposed method can indeed cope with the all-day (\emph{i.e.}, daytime and nighttime) scene flow estimation, but usually fails for adverse weather, such as rain and fog. Rain will lead to varying degrees of dynamic streak noise to RGB, event and LiDAR, and fog will attenuate atmospheric light, limiting the imaging of these sensors. In the future, we will further introduce the near-infrared sensor into these modalities to handle adverse weather scene flow.

\section{Conclusion}
In this work, we bring the event as a bridge between RGB and LiDAR, and propose a novel hierarchical visual-motion fusion framework to fuse the multimodal complementary knowledge in the homogeneous space for scene flow. To the best of our knowledge, we are the first to investigate the inter-modal homogeneous space to explicitly learn the complementary knowledge for physical interpretation. We construct the Event-RGB luminance space, Event-LiDAR structure space and RGB-Event-LiDAR correlation space. The three homogeneous spaces progressively improve scene flow from visual space to motion space. We also conduct extensive experiments on daytime and nighttime scenes, and verify the superiority of the proposed method. We believe that the proposed method could not only facilitate the development of scene flow but also enlighten the researchers of the broader field, \emph{e.g.}, adverse scene multimodal perception.

\noindent
\textbf{Acknowledgments.} The computation is completed in the HPC Platform of Huazhong University of Science and Technology. This work was supported in part by the National Natural Science Foundation of China under Grant 62371203.

{
  \small
  \bibliographystyle{ieeenat_fullname}
  \bibliography{egbib}
}

\end{document}